\documentclass{article}

\usepackage{PRIMEarxiv}

\usepackage[utf8]{inputenc} 
\usepackage[T1]{fontenc}    
\usepackage{hyperref}       
\usepackage{url}            
\usepackage{booktabs}       
\usepackage{amsfonts}       
\usepackage{nicefrac}       
\usepackage{microtype}      
\usepackage{lipsum}
\usepackage{textcomp}
\usepackage{fancyhdr}       
\usepackage{graphicx}       
\usepackage{algorithm}
\usepackage{amsmath}
\usepackage{comment}
\usepackage{algpseudocode}
\usepackage{placeins}
\usepackage{tabularx}
\usepackage{subcaption}
\usepackage{amssymb}
\graphicspath{{media/}}     
\usepackage[absolute,overlay]{textpos}
\title{FastODT: A tree-based framework for efficient continual learning
}

\author{
Daniel Bretsko\textsuperscript{*} \\
Masaryk University \\
Brno, Czech Republic \\
\texttt{bretsko@math.muni.cz}
\And
Piotr Walas\textsuperscript{*} \\
Masaryk University, Brno, Czech Republic \\
Warsaw University of Technology, Warsaw, Poland\\
\texttt{walas.piotr@outlook.com}
\And
Devashish Khulbe\textsuperscript{*} \\
Masaryk University \\
Brno, Czech Republic \\
\texttt{khulbe@math.muni.cz}
\And
\And
Sebastian Stros \\
Mycroft Mind \\
Brno, Czech Republic \\
\texttt{sebastian.stros@mycroftmind.com}
\And
Stanislav Sobolevsky \\
Masaryk University, Brno, Czech Republic \\
New York University, New York, NY, USA \\
Analog AI, Abu Dhabi, UAE \\
\texttt{ss9872@nyu.edu}
\And
Tomas Satura \\
Mycroft Mind \\
Brno, Czech Republic \\
\texttt{tomas.satura@mycroftmind.com}
}

\begin{document}
\maketitle
\renewcommand{\thefootnote}{\fnsymbol{footnote}}
\footnotetext[1]{Equal contribution}

\begin{abstract}

Machine learning models deployed in real-world settings must operate under evolving data distributions and constrained computational resources. This challenge is particularly acute in non-stationary domains such as energy time series, weather monitoring, and environmental sensing. To remain effective, models must support adaptability, continuous learning, and long-term knowledge retention. This paper introduces a oblivious tree-based model with Hoeffding bound controlling its growth. It seamlessly integrates rapid learning and inference with efficient memory management and robust knowledge preservation, thus allowing for online learning. Extensive experiments across energy and environmental sensing time-series benchmarks demonstrate that the proposed framework achieves performance competitive with, and in several cases surpassing, existing online and batch learning methods, while maintaining superior computational efficiency. Collectively, these results demonstrate that the proposed approach fulfills the core objectives of adaptability, continual updating, and efficient retraining without full model retraining. The framework provides a scalable and resource-aware foundation for deployment in real-world non-stationary environments where resources are constrained and sustained adaptation is essential.
\end{abstract}

\keywords{Continual Learning \and Sequential data modeling \and Tree-based models \and Hoeffding bound \and Obliviousness}

\section{Introduction}
\subsection{Background}
Continual and incremental learning (IL) remains a key frontier in machine learning, enabling models to adapt to evolving data distributions without requiring complete retraining. This capability is essential in real-world settings where data arrives sequentially or the underlying data distribution drifts over time. Incremental learning supports efficient adaptation and long-term retention, making it particularly valuable in domains characterized by non-stationary and streaming data, such as energy demand forecasting, real-time traffic prediction, weather monitoring, and environmental sensing \cite{lu2019lifelong}. Beyond these traditional applications, continual learning has recently gained prominence in large-scale foundational
models, where the goal is to incrementally incorporate new knowledge while preventing catastrophic forgetting \cite{parisi2019continual,ke2024continual}. Incremental fine-tuning and online adaptation strategies are researched to keep ML models up to date with evolving facts, domains, and user preferences without costly retraining on massive corpora. Broadly, supervised incremental learning algorithms can be categorized into three main scenarios: Task-incremental, Domain-incremental, and Class-incremental learning \cite{vandeVenTuytelaarsTolias2022}. In this work, we focus on Domain-incremental learning methods, specifically on the short-term time-series forecasting task, where the predictive model adapts to conceptually drifted distributions of lagged target variables.

\subsection{Related Work}
Although most high-performing approaches to incremental learning are combinations of past approaches, there are at least two relevant ways to categorize IL approaches. The first categorization states that IL algorithms in the non-neural network part of machine learning can be divided into tree-based and non-tree-based approaches. Tree-based IL extends decision and regression trees to streaming contexts by implementing an update mechanism into how the trees are constructed. The most famous example is the Very Fast Decision Tree (VFDT) or Hoeffding Tree \cite{domingos2000vfdt}. Unlike conventional approaches to splitting, which require full data, VFDT and its offshoots use the Hoeffding bound to determine statistically reliable splits from an incomplete dataset. In comparison, non-tree-based approaches have an updating mechanism implemented on a higher level than that of tree growing. This could involve tree deletion, weighting, instance weighting, forgetting, or even retraining the entire model. There are so many non-tree-based approaches that a second more descriptive categorization approach is needed.

The second categorization approach introduces three categories \textit{windowing}, \textit{performance-based methods}, and \textit{ensemble methods}. \textit{Windowing} selects and weights instances into a window (or windows) on which the ML model is retrained. One of the oldest algorithms using this approach is STAGGER \cite{STAGGER}. Currently, the most popular algorithm that retrains on a dynamically expanding and contracting window of the most relevant instances is ADWIN \cite{bifet2007learning}. In contrast to picking the window of the most relevant instances irrespective of model performance, \textit{performance-based methods} initiate retraining of the machine learning model when its accuracy drops or significantly deteriorates. State of the art algorithm of the approach is the Drift Detection Method (DDM), which treats deteriorated model performance as a rejection of the null hypothesis that the current data distribution has not changed \cite{DDM_OG}. The model's bad performance is a temporal threshold from which point on window of new data is collected. After the performance worsens by 3$\sigma$, the model is retrained on the data collected since the temporal threshold. Although both performance-based and windowing approaches generated burgeoning literature from the 1990s until the 2010s, the focus and highest performance on conceptually drifted data has shifted toward the ensemble methods. Our approach combines tree-based and ensemble methods.

While building on previous incremental learning approaches and static machine learning ensemble models, incrementally learning ensemble methods introduce tree removal, retraining, and weighting. Ensemble methods such as Online Bagging \cite{oza2005online}, Learn++.NSE \cite{polikar2001learn++}, and Adaptive Random Forest (ARF) \cite{gomes2017arf} combine dynamic tree replacement, drift detection, and instance weighting to maintain accuracy under concept drift. More recently, incremental gradient boosting variants have emerged, aiming to retain the predictive power of gradient-boosted ensembles in online environments. iGBDT \cite{zhang2019igbdt} incrementally updates gradient boosting trees by merging new batches with existing data, selectively rebuilding sub-tree structures when split attributes change without explicit drift detection thresholds. In contrast, eGBDT \cite{wang2023egbdt}, and GBDT-IL \cite{chen2024gbdtil} extend the framework with pruning and rollback mechanisms that remove or revert under-performing trees when drift occurs, using ensemble-level residual monitoring and, in the case of GBDT-IL, Fisher Score–based feature selection. These methods improve the balance between adaptability and stability by integrating additive and subtractive updates, though they still face computational challenges due to the sequential nature of boosting. Overall, tree-based incremental ensemble methods offer interpretable, modular, and resource-efficient solutions, with ARF and incremental GBDTs representing two complementary directions, adaptive bagging versus boosting, for concept-drift–aware learning. In contrast, non-tree-based IL methods encompass neural, kernel, and probabilistic models—such as incremental SVMs \cite{cauwenberghs2001svm}, evolving neural networks \cite{lee2017den}, Bayesian learners \cite{kochurov2018bayesian}, and prototype-based models \cite{nova2014lvq}, which incrementally update parameters, support vectors, or prototypes. These methods often achieve strong flexibility and generalization but at the cost of higher computational demands and reduced interpretability, making tree-based approaches particularly attractive for real-time, large-scale streaming applications.

\subsection{System Proposal}

Our work proposes FastODT, a novel incremental learning framework that combines the efficiency of Oblivious Decision Trees (ODTs) with the adaptive growth of VFDTs. FastODT employs the Hoeffding Bound to incrementally expand the tree structure while maintaining a compact ODT-like form that eliminates traversal during inference. This results in low memory usage and high-speed predictions. However, once a tree reaches its maximum depth, learning halts to prevent an infinite increase in memory demand. End of growing makes the model susceptible to concept drift, which can be prevented by employing FastODTs in ensemble method. Particularily, two ensemble strategies are explored: IncubationBoost, a boosting-based extension of elastic GBDT where new trees grow in parallel and “hatch” to replace underperforming trees, and a bagging-based variant of Adaptive Random Forests (ARF) with built-in drift detection for dynamic tree replacement. Together, the two strategies make FastODT an efficient, drift-resilient online learning model suitable for evolving data streams.

We focus our experiments on highly granular time series data, including high-frequency household and commercial energy consumption and air quality datasets. Such datasets are characterized by continuous streams and rapid temporal dynamics, which require predictive models capable of efficient online adaptation and real-time inference. These datasets allow us to evaluate our methods for domain-incremental models by learning the same target variable under changing distributions and input feature contexts. Tree-based methods are particularly well-suited for these scenarios due to their computational efficiency, interpretability, and robustness to non-stationary data \cite{domingos2000vfdt, manapragada2018extremely}. Moreover, in practical applications such as smart meters, sensor networks, and edge-based environmental monitoring, models must operate under strict memory and compute constraints \cite{gama2014survey, bifet2010moa}. These conditions make lightweight, incrementally updatable tree algorithms a compelling choice, enabling fast, adaptive learning without requiring full data storage or retraining.

Our experiments establish the efficiency and accuracy of FastODT, a practical, resource-conscious incremental learning solution tailored for real-world sensor, energy, and environmental monitoring pipelines, where rapid adaptation and low computational overhead are essential. In summary, this work makes the following key contributions: 
\begin{itemize}
    \item \textit{FastODT}, a new incremental tree-learning method that merges the compact structure of Oblivious Decision Trees with the statistical growth guarantees of VFDTs, enabling fast, memory-efficient updates without requiring full data revisitation.
    \item \textit{IncubationBoost}, A drift-aware ensemble framework that leverages FastODT as a base learner, a boosting-based ensemble that incrementally grows and “hatches" new trees to replace degraded ones without explicit drift thresholds.
\end{itemize}


\section{Algorithms}
\label{sec:methods}

\subsection{Oblivious Decision Trees (ODTs)}

\begin{figure}[htbp]
    \centering
    
    \begin{subfigure}{0.48\linewidth}
        \centering
        \includegraphics[height=6cm]{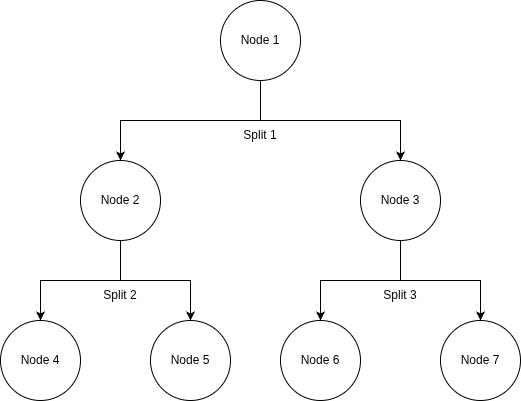}
        \caption{Standard Decision Tree}
        \label{fig:DT}
    \end{subfigure}
    \hfill
    \begin{subfigure}{0.48\linewidth}
        \centering
        \includegraphics[height=6cm]{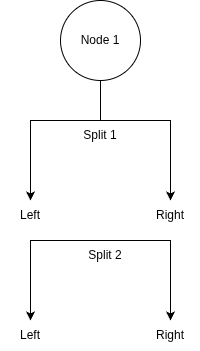}
        \caption{Oblivious Decision Tree}
        \label{fig:ODT}
    \end{subfigure}
    
    \caption{Comparison of standard Decision Tree and Oblivious Decision Tree structures.}
    \label{fig:DT_vs_ODT}
\end{figure}

Oblivious Decision Trees (ODTs) are a constrained form of decision trees in which all nodes at the same depth split on the same feature and threshold. In standard Decision Trees, each split leads to a new node, and each node has its own split except for the final nodes, which serve as output or final value provider, as displayed in Figure \ref{fig:DT}. This architecture leads to the need for tree traversal, meaning going through nodes and checking where splits lead. 

Oblivious Decision Trees implement depth-wise common splits, i.e., the same split for all nodes at the same tree depth. If the tree in the Figure \ref{fig:DT} were ODT, then Split 1 and Split 2 would be basically the same split, having the same threshold value and being shared between all the nodes at the same depth. ODTs are minimizing splits into a series of left and right decisions where particular nodes stop playing a significant role, hence they can be described by Figure \ref{fig:ODT}.

This structure produces a compact and memory-efficient model, as each tree level applies a uniform rule to all samples. Consequently, ODTs can avoid tree traversal during inference, due to known splits ahead of time, making them very computationally effective with complexity fit $O(n)$ where $n$ is the number of data samples, contrary to the standard decision tree, which has complexity $O(d*n)$, where $d$ is the tree depth. The homogeneous splitting structure helps to reduce overfitting by limiting split flexibility, at the cost of potentially reduced representational power compared to standard decision trees. This trade-off can be mitigated through deeper trees or ensembles. ODTs are particularly well-suited for resource-constrained environments, such as embedded or edge devices, where computational and memory resources are limited, and rapid predictions are required.

Although ODTs offer exceptional efficiency during inference, they suffer from retraining in online scenarios or streaming scenarios. ODTs cannot update their states based on newly arrived samples without a full retraining procedure.

\subsection{Very Fast Decision Trees with Hoeffding Bound (VFDT)}

Alternatively, the retraining problem is addressed by Very Fast Decision Trees with Hoeffding Bound. VFDTs are specifically designed to work with incoming streaming or online data by training on each arriving sample and building a set of statistical trackers that compress historical data and represent the dataset information over time. These statistical trackers are not lossless, so the tree needs to infer when to split based on somewhat limited information rather than the full dataset. This is crucial, as streaming data can be perceived as infinite in size, with the tree only able to observe a chunk at a time. This way, the amount of stored data required for the tree would have to be windowed, or it could be enormous, making computation infeasible. In VFDTs, this is solved with the so-called Hoeffding bound, which provides a dynamic threshold for the tree, with statistical guarantees for determining if sufficient evidence exists to perform a node split. Given the confidence parameter, it estimates the number of samples required to differentiate between the best and the second-best splits, where each split is defined by a feature and a threshold.

\begin{figure}
    \centering
    \includegraphics[width=0.5\linewidth]{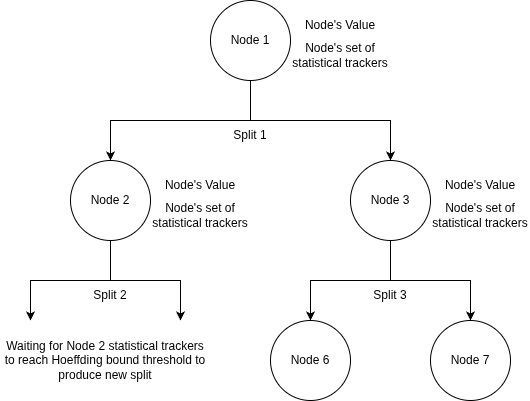}
    \caption{Very Fast Decision Tree}
    \label{fig:VFDT}
\end{figure}

As presented in Figure \ref{fig:VFDT}, these statistical trackers and the Hoeffding bound are stored and computed node-wise, hence each node has a different snapshot of the dataset depending on which branch of the tree it is and how often it is updated with incoming data. After the split, these statistical trackers stop being useful and can be discarded. Whether they are discarded depends on the implementation's choices.

The Hoeffding tree reduces training and retraining to a single pass through the entire dataset even in case when dataset is not fully observable yet. That is a major difference from the traditional Decision Trees, which have to perform a pass each time new data arrives when considering online or dynamic datasets. That means a conventional Decision Tree has to be retrained each time the dataset changes. VFDT has flexibility of just appending new data sample to its knowledge-base, by upadating statistical trackers. This process can be considered as single pass even when not all data is known at the time, as VFDT never has to visit past data samples.

\subsection{FastODT: Incremental Learning Adaptation}

The proposed Fast Oblivious Decision Tree (FastODT) combines elements from both solutions, providing a tree with fast inference, reduced memory usage, and one-pass learning for streaming and online data. FastODT, similarly to VFDT, needs only one dataset pass even while the dataset is evolving. This is achieved by the implementation of the Hoeffding bound to determine statistically confident split decisions without revisiting historical data. In addition, FastODT uses a single split per depth from the ODT design, leading to a smaller memory footprint than VFDT, since only one split is implemented per depth. 

\begin{figure}
    \centering
    \includegraphics[width=0.5\linewidth]{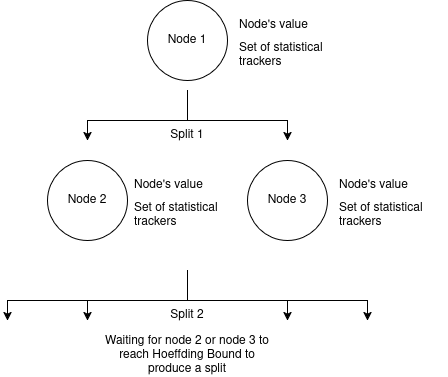}
    \caption{Fast ODT}
    \label{fig:FastODT}
\end{figure}

As per Figure \ref{fig:FastODT}, in FastODT, each node has its own set of statistical trackers, similar to VFDT, however, as only one split can exist per depth due to obliviousness, nodes are competing for reaching the split first. Whichever node produces a new split as the first wins. That split is then used for that depth indefinitely. After that, the process is repeated for the next depth, with statistical trackers from the previous depth rejected because they are no longer useful. In practical implementations, it is more likely that FastODT will not be a fully balanced tree. This means that the least-visited branches will not be as deep as the most-visited ones. In such case, instead of rejecting statistical trackers depth-wise, the same process can be implemented by storing trackers only in the bottom nodes branch-wise. 

As a result, FastODT achieves both the low-latency inference of ODTs and the adaptivity of VFDTs, making it highly suitable for evolving data streams where full data storage and retraining are infeasible.

\begin{algorithm}
\caption{FastODT Learning with node Traversal}
\begin{algorithmic}[1]
\Procedure{FastODTnode}{currentnode, sample, labels, depth}

\If{Splits[depth] exists}
    \If{sample < Splits[depth]}
        \State \Call{FastODTnode}{currentnode.Leftnode, sample, labels, depth + 1}
    \Else
        \State \Call{FastODTnode}{currentnode.Rightnode, sample, labels, depth + 1}
    \EndIf

\Else
    \State hf $\gets$ \Call{HoeffdingBound}{currentnode, sample}

    \If{hf > split\_threshold}
        \State currentnode.Leftnode $\gets$ new FastODTnode
        \State currentnode.Rightnode $\gets$ new FastODTnode
        \State Splits[depth] $\gets$ \Call{ComputeBestSplit}{currentnode}
    \Else
        \State \Return
    \EndIf
\EndIf

\EndProcedure
\end{algorithmic}
\end{algorithm}

\begin{algorithm}
\caption{FastODT Learning with Binary Mask Indexing}
\begin{algorithmic}[1]
\Procedure{FastODTUpdate}{nodes, sample, labels, Splits}
    \State depth $\gets$ $|$Splits$|$ \Comment{Current tree depth}
    \State binaryMask $\gets$ $[$ sample $<$ Splits[i] for $i = 0 \dots$ depth $- 1$ $]$
    \State nodeIndex $\gets$ $\sum_{i=0}^{\text{depth}-1}$ binaryMask[i] $\cdot 2^i$ \Comment{Convert binary to integer}
    \State currentnode $\gets$ nodes[depth][nodeIndex]
    \State hf $\gets$ \Call{HoeffdingBound}{currentnode, sample, labels}
    \If{hf $>$ split\_threshold}
        \State splitValue $\gets$ \Call{ComputeBestSplit}{currentnode}
        \State Splits[depth] $\gets$ splitValue
        \State nodes[depth$+1$][$2 \cdot$ nodeIndex] $\gets$ new FastODTnode
        \State nodes[depth$+1$][$2 \cdot$ nodeIndex $+ 1$] $\gets$ new FastODTnode
    \Else
        \State \Return
    \EndIf
\EndProcedure
\end{algorithmic}
\end{algorithm}

\FloatBarrier
\subsubsection{Adaptive Histograms for Split Selection}

To implement the Hoeffding-based splitting mechanism efficiently, FastODT employs adaptive histograms to approximate feature distributions in a streaming setting. Unlike traditional batch histograms used in gradient boosting frameworks (e.g., XGBoost), these histograms are updated incrementally as new samples arrive. When the Hoeffding bound criterion is met for a feature, the best-scoring histogram bin determines the optimal split threshold. The granularity of these histograms (i.e. the number of bins) can be dynamically adjusted, allowing control over memory usage and split precision. This adaptability enables FastODT to balance model compactness and accuracy depending on resource constraints and data variability.

\FloatBarrier
\subsubsection{Online Boosting Algorithm - Incubation Boost}
With concept drift in the dataset and real-time updates, classic boosting algorithms must constantly train new trees on already collected data, causing significant computational overhead.

Because FastODT can be trained in an online fashion, it provides the opportunity to maintain trees alongside the boosting algorithm allowing them to learn incrementally as new data arrives, without imposing a heavy computational workload. Furthermore, FastODT can naturally determine when it has been trained on sufficient data by reporting the number of splits it has generated. If the number of splits matches the expected depth of the tree, this indicates that the tree has reached readiness and can be incorporated into the boosting algorithm.

To implement this scheme, we propose a new tree boosting algorithm specifically designed for FastODT, called Incubation Boost. Incubation Boost works similarly to eGBDT boosting algorithm. The key difference is that new trees are not immediately added to the boost chain. Instead, a tree enters an incubation period during which it learns from incoming data. The tree remains in incubation until it reaches its expected depth. Once this condition is met, it is added to the ensemble; this process is reffered to as hatching. After a tree is hatched and added to the chain, the next tree enters the incubation period until it is also hatched.

All trees go through an incubation period, with a slight exception for the very first tree in the Incubation Boost algorithm. Unlike traditional boosting methods, Incubation Boost does not start with a fully formed ensemble; instead, the chain grows incrementally until the expected number of trees is reached. Therefore, Incubation Boost may start with zero trees, with the first tree being initialized upon the arrival of training data.

Additionally, the initial data does not guarantee that the first tree will leave the incubation process immediately. This creates an edge case in which the first tree is still incubating, leaving Incubation Boost unable to provide predictions. Since the main goal is to keep Incubation Boost operational throughout its entire lifetime, this situation must be avoided. Therefore, the first tree contributes to predictions even while it is still in the incubation process.

Starting from the second tree onward, incubating trees do not participate in the prediction process until they are fully hatched.

As the Incubation Boost chain grows towards the expected number of trees, it must incorporate pruning techniques to adapt to concept drift. By default, this is handled by monitoring the error rate of each tree and pruning those with the worst performance.

The entire process of training\ref{alg:incubation-training} and inference\ref{alg:incubation-inference} is presented in the form of pseudocode.

\begin{algorithm}
\caption{Incubation Boost -- Online Update for One Sample}
\label{alg:incubation-training}
\begin{algorithmic}[1]
\Require Input sample $(\mathbf{x}, y)$
\State $r \gets y$
\State $\hat{y} \gets 0$
\State $b \gets \varnothing$
\State $\epsilon_{\min} \gets \infty$

\If{ensemble is empty}
    \State Initialize new tree $T$
    \State Append $T$ to ensemble
\EndIf

\For{each tree $T_i$ in ensemble with index $i$}
    \State $\hat{y}_i \gets \eta \cdot T_i(\mathbf{x})$
    \State $\hat{y} \gets \hat{y} + \hat{y}_i$
    \State $r \gets y - \hat{y}$

    \If{$T_i$ is fully grown}
        \State $\epsilon \gets \textsc{ComputeError}(T_i, y, \hat{y})$
        \If{$\epsilon < \epsilon_{\min}$}
            \State $b \gets i$
            \State $\epsilon_{\min} \gets \epsilon$
        \Else
            \State \textbf{break}
        \EndIf
    \Else
        \Comment{Update incubating tree}
        \State Update $T_i$ with $(\mathbf{x}, r)$
    \EndIf
\EndFor

\If{$b \neq \varnothing$}
    \State Remove trees with index $> b$
\EndIf

\State $T_{\text{new}} \gets \textsc{Incubate}(\mathbf{x}, r)$

\If{$T_{\text{new}} \neq \varnothing$ \textbf{and} ensemble size $< L$}
    \State Append $T_{\text{new}}$ to ensemble
\EndIf

\end{algorithmic}
\end{algorithm}

\begin{algorithm}
\caption{Incubation Boost -- Prediction}
\label{alg:incubation-inference}
\begin{algorithmic}[1]
\Require Input feature vector $\mathbf{x}$
\State $\texttt{ensemble\_prediction} \gets 0$

\For{each $\texttt{tree}_i$ in ensemble (index $i$)}
    \State $\texttt{weighted\_prediction} \gets \texttt{learning\_rate} \cdot \texttt{tree}_i(\mathbf{x})$
    \State $\texttt{ensemble\_prediction} \gets 
           \texttt{ensemble\_prediction} + \texttt{weighted\_prediction}$

\EndFor

\State \Return $\texttt{ensemble\_prediction}$

\end{algorithmic}
\end{algorithm}

\FloatBarrier
\section{Experiments and Results}

\subsection{Data and Experiment setup}

For evaluation, we mainly use high-resolution temporal datasets. These are derived from proprietary energy consumption (industrial and commercial) based anonymized smart meters from MycroftMind\textsuperscript{TM} (MM). Both commercial and industrial datasets consist of historical records from multiple smart meters corresponding to different sites or locations. The commercial data contains records from 3 unique sites, whereas the industrial data was in two different "batches", the first one contained data from 5 devices (smart energy meters), whereas the second batch had data from 7 sites. 

Another energy consumption dataset consists of historical consumption data from a household in France and is publicly available as an open-source dataset \cite{individual_household_electric_power_consumption}. We also focus on granular Air Quality (AQ) data \cite{uci2015pm25}, which include high-resolution measurements of various physical AQ indicators, such as PM2.5, along with weather measurements (temperature, pressure, etc.). Additional considered dataset is the Friedman data, which is a synthetically generated regression benchmark originally introduced by Friedman for evaluating nonlinear regression models. In its standard formulation (Friedman~\#1), the target variable is defined as

\begin{equation}
\label{friedmanEq}
    y = 10 \sin(\pi x_1 x_2) + 20 (x_3 - 0.5)^2 + 10 x_4 + 5 x_5 + \epsilon
\end{equation}

where $x_i \sim \mathcal{U}(0,1)$ for $i = 1, \dots, 10$, and $\epsilon \sim \mathcal{N}(0, \sigma^2)$ represents additive Gaussian noise. Only the first five features are informative, while the remaining features are irrelevant noise variables. 

\begin{table*}[h]
\centering
\caption{Overview of datasets used for experiments}
\label{tab:data_stats}
\begin{tabularx}{\textwidth}{l c l c l}
\toprule
\textbf{Data} & \textbf{\# Samples} & \textbf{Target} & \textbf{Granularity} & \textbf{Timeline} \\
\midrule
Air Quality & 43,800 & PM2.5 & 1-hour & 2010--2014 \\

Household electricity & 2,075,259 & Active Power (kW) & 1-minute & 2006--2010 \\

Friedman & 10,000 & $y$\textsuperscript{~\ref{friedmanEq}} & NA & NA \\

Industrial power (MM-1st batch) 
& 19,038,807 & Power (kW) & 1--2 second & 09-2022--05-2023 \\

Industrial power (MM-2nd batch) 
& 361,166 & Total Energy (kWh) & 15-minute & 02-2023--04-2025 \\

Commercial power (MM-2nd batch) 
& 24,362 & Total Energy (kWh) & 5-minute & 06-2024--08-2024 \\
\bottomrule
\end{tabularx}
\end{table*}

The Friedman data has been widely used as an experimentation benchmark for incremental and data stream learning models \cite{ikonomovska2009regression}, as it provides controlled nonlinearity, feature interactions, and noise characteristics suitable for evaluating adaptive regression algorithms. 


We aggregate the target features in all our datasets on an hourly basis to remove noisy components and make practical prediction models for real-world use cases. All datasets span a sufficiently long time to incorporate seasonality and cyclical trends in the prediction models. Table~\ref{tab:data_stats} shows the statistics for each of the data involved in the experiments.



All models are trained to predict the one-step residual $r_t = y_t - y_{t-1}$, 
i.e., $\hat{r}_t = f_\theta(\mathbf{x}_t)$, and the final forecast is obtained as 
$\hat{y}_t = y_{t-1} + \hat{r}_t$.
\subsection{Results}
\begin{table*}[h]
\centering
\caption{MAPE (\%) comparison across datasets}
\label{tab:mape_results}
\begin{tabular}{lcccccc}
\hline
\textbf{Model} & \textbf{Air Quality} & \textbf{Friedman} & \textbf{Electricity} & \textbf{Commercial} & \textbf{Industrial(1)} & \textbf{Industrial(2)}\\
\hline
Baseline (Mean value) & 21.741 & 8.429 & 48.709 & 2.127 & 22.004 & 8.656 \\
GBDT                  & 20.241 & 8.339 & 43.726 & 3.623 & 17.972 & 6.122 \\
Random Forest         & 20.545 & 8.330 & 46.124 & 3.091 & 15.932 & 6.020 \\
Hoeffding Tree        & 20.371 & 8.331 & 45.314 & 1.926 & 21.456 & 7.043 \\
ARF (Hoeffding)        & 20.959 & 8.337 & 46.671 & 1.853 & 16.294 & 5.799 \\
FastODT               & 20.254 & 8.344 & 44.619 & 3.076 & 22.189 & 7.115 \\
ARF (FastODT)          & 20.294 & 8.341 & 49.727 & 2.542 & 21.087 & 6.478 \\
IncubationBoost       & 20.458 & 8.344 & 45.591 & 1.963 & 16.070  & 6.237 \\
\hline
\end{tabular}
\end{table*}

\begin{table*}[h]
\centering
\caption{RMSE comparison across datasets}
\label{tab:rmse_results}
\begin{tabular}{lcccccc}
\hline
\textbf{Model} & \textbf{Air Quality} & \textbf{Friedman} & \textbf{Electricity} & \textbf{Industrial(1)} & \textbf{Industrial(2)} & \textbf{Commercial} \\
\hline
Baseline (Mean value) & 25.787 & 1.528 & 0.554 & 0.190 & 1.449 & 0.066 \\
GBDT                  & 25.755 & 1.515 & 0.572 & 0.173 & 1.095 & 0.049 \\
Random Forest         & 25.291 & 1.513 & 0.527 & 0.167 & 1.070 & 0.051 \\
Hoeffding Tree        & 25.486 & 1.515 & 0.591 & 0.198 & 1.264 & 0.055 \\
ARF (Hoeffding)       & 25.362 & 1.514 & 0.565 & 0.186 & 1.073 & 0.058 \\
FastODT               & 25.901 & 1.516 & 0.592 & 0.212 & 1.261 & 0.070 \\
ARF (FastODT)         & 25.828 & 1.516 & 0.572 & 0.205 & 1.116 & 0.070 \\
IncubationBoost       & 25.659 & 1.516 & 0.583 & 0.188 & 1.058 & 0.056 \\
\hline
\end{tabular}
\end{table*}

Table~\ref{tab:mape_results} and Table~\ref{tab:rmse_results} show the Mean Absolute Percentage Error (MAPE) values and Root Mean Square Error (RMSE) Values across models and datasets. For Industrial and Commercial data from MycroftMind, results for only one device/site are shown for brevity. In the Air Quality dataset, all methods exhibit relatively close performance, with MAPE values concentrated around 20\%. The standard GBDT achieves the lowest error, closely followed by FastODT. The proposed FastODT ensembles show competitive performance, with ARF, and IncubationBoost remaining within a narrow margin of the best-performing model. Classical incremental learners such as Hoeffding Tree and ARF (Hoeffding) perform slightly worse, indicating limited gains from incremental bagging on this relatively stable stream. For the Friedman dataset, all models again demonstrate nearly identical performance. The best results are obtained by the Random Forest and Hoeffding Tree models, while GBDT follows closely. FastODT and its ensemble variants (FastODT, ARF FastODT, and IncubationBoost) preserve accuracy relative to strong batch and incremental baselines in this synthetic regression setting. In contrast, larger performance differences emerge on the Electricity dataset, which exhibits stronger non-stationarity. Here, the batch GBDT outperforms all incremental learning model variants. Among incremental models, IncubationBoost further reduces error compared to classical Hoeffding-based ensembles, demonstrating the benefit of the proposed boosting-style replacement strategy under drift. However, the ARF-based FastODT variant shows degraded performance, suggesting that naive bagging of shallow oblivious trees may be less suitable in highly volatile electricity demand streams. 

Across the industrial datasets, ensemble methods consistently outperform the baseline and single-tree models. For Industrial(1), Random Forest provides the most stable and accurate performance under both RMSE and MAPE, while boosting and adaptive forest variants remain competitive. On Industrial(2), enhanced boosting and adaptive forest approaches achieve the strongest results, indicating the benefit of ensemble adaptation and pruning strategies for more complex industrial dynamics. In the Commercial dataset, Gradient Boosting achieves the best performance in terms of RMSE, whereas adaptive forest methods perform better under MAPE. Overall, ensemble-based models dominate across industrial and commercial settings, with performance varying slightly depending on whether absolute or relative error is considered.

Overall, the results show that FastODT maintains accuracy on par with established batch and incremental tree methods on stable and moderately drifting streams, while its boosting-based ensemble variants provide improved robustness under stronger concept drift. Importantly, these competitive error rates are achieved alongside the computational and memory advantages of the proposed FastODT framework, reinforcing its suitability for real-time, resource-constrained streaming applications.

\subsection{Discussion}
While FastODT is a novel online learning approach with strong computational efficiency and memory efficiency and achieves competitive accuracy on the tested datasets, several limitations are worth considering.

Firstly, the obliviousness is a strong form of regularization which prevents overfitting but can introduce too much bias. Single split per depth reduces variance and memory footprint but restricts representational flexibility. In highly heterogeneous feature spaces, where different regions of the input distribution require different split features at the same depth, FastODT may underfit and thus perform worse than fully adaptive node-wise trees. This limitation becomes more pronounced in high-dimensional problems.

Secondly, split decisions are irreversible. Once a depth-wise split is committed, FastODT does not revise or replace it. Although ensemble-level pruning mitigates performance degradation under drift, individual tree structures remain static after reaching maximum depth. Arguably, the ensemble-level pruning may not suffice to perform well on datasets with rapidly changing concept drift. 

Thirdly, the method relies on histogram-based split approximation. The coarseness of the histogram binning is a important parameter which might not be trivial to tune. Coarser granularity of binning is more memory-efficient and introduces regularization. On the other hand, fine binning increases memory and update cost, but will likely lead to more precise albeit potentially overfitting splits. Thus, the parameter not only introduces bias-variance trade-off, but also change memory demands.

Finally, the FastODT is so far only tested in few ensemble models and only on some datasets. Great benefit of the model is that it is compatible with many different ensembles; however, that remains to be tested. Equally, more robust testing in terms of datasets could tell us more about its performance in different contexts.

\section{Conclusion}

In this work, we introduced \textit{FastODT}, a novel incremental tree learning algorithm that bridges the computational and memory efficiencies of Oblivious Decision Trees with the statistically grounded growth mechanism of Hoeffding Trees. By enforcing depth-wise shared splits while leveraging the Hoeffding bound for split confidence, FastODT achieves one-pass learning without revisiting historical data, while preserving the low-latency inference and compact memory footprint characteristic of oblivious trees. This design directly addresses the limitations of conventional ODTs in streaming settings and the higher memory demands of fully node-wise VFDT-style trees.

Empirical evaluation on high-resolution energy consumption, air quality, and synthetic regression benchmarks demonstrate that FastODT achieves predictive performance comparable to strong batch-learning baseline models and established incremental methods. On stable and moderately drifting datasets, FastODT preserves accuracy while benefiting from reduced computational overhead. Under stronger non-stationarity, the proposed boosting-based ensemble variants improve robustness relative to classical Hoeffding-treee-based ensembles. Importantly, these results are obtained without full retraining or complete data storage, highlighting the practical relevance of the approach for real-world streaming scenarios.

Additionally, to test FastODT's online ensemble learning features, we have adapted a boosting strategy by introducing Incubation Boost, an elastic boosting algorithm designed for concept drift prevention and online learning, which has proven to provide competitive results. Incubation Boost introduces a novel strategy for adding trees to the boosting chain by training them in an incubation–hatch process, allowing the model to avoid windowing and storing training data.

Overall, FastODT provides a lightweight, interpretable, and resource-efficient solution for domain-incremental time-series forecasting. Its compact structure, single-pass updates, and ensemble compatibility make it particularly suitable for edge computing environments, smart meters, and sensor networks where memory and latency constraints are critical. Future work may extend FastODT towards deeper adaptive split revision, hybrid boosting–bagging combinations, and integration with drift-aware feature selection or representation learning techniques, further strengthening its applicability to large-scale incrementally learning systems.

\clearpage
\thispagestyle{empty}

\begin{textblock*}{5cm}(1.5cm,26.5cm)
  \includegraphics[width=2.6cm]{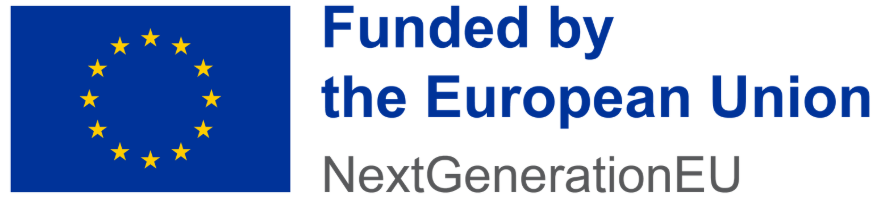}
\end{textblock*}

\begin{textblock*}{5cm}(16.5cm,26.5cm)
  \includegraphics[width=3.6cm]{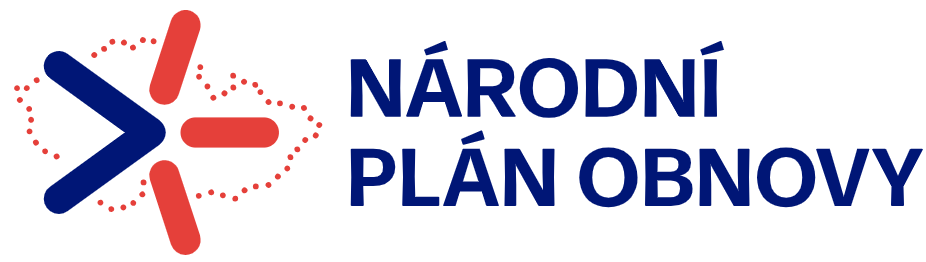}
\end{textblock*}

\section*{Acknowledgments}
This paper is written as part of project No. 2320000007 named Self-learning Autonomous AIoT Devices (SLAID), financed from the National Recovery Plan, subcomponent 1.5.1.4 Support for IPCEI in Microelectronics and Communication Technology.







\end{document}